\documentclass[conference]{IEEEtran}
\IEEEoverridecommandlockouts
\usepackage{vector} % local vector.sty (fixed boldface greek math)
\usepackage{tikz}
\usepackage{algorithmicx}
\usepackage{algorithm}
\usepackage{algpseudocode}
\usepackage{amsmath}
\usepackage{amsfonts}
\usepackage{siunitx}
\usepackage{commath}
\usepackage[backend=biber,style=ieee,natbib=true]{biblatex} 
\usepackage{lipsum}
\usepackage{multirow}
\usepackage{bbm}
\usepackage{hyperref}
\usepackage{cleveref}
\usepackage[keeplastbox]{flushend}

\usepackage{amsmath} % JIC.
\usepackage{algorithmicx}
\usepackage{algorithm}
\usepackage{algpseudocode}
\usepackage{amsmath}
\usepackage{amsfonts}
\usepackage{siunitx}
\usepackage{commath}

\usepackage{vector} % JIC. (local vector.sty)

\DeclareMathOperator{\Geo}{Geo}

\newcommand{\E}{\mathbb{E}}

\newcommand{\w}{\mathbf{w}}

\newcommand{\M}{\mathbf{M}}
\newcommand{\m}{\mathbf{m}}

\newcommand{\e}{\mathbf{e}}
\newcommand{\bfE}{\mathbf{E}}
\newcommand{\surrogate}{\hat{S}} % Surrogate model (\mathcal{S})
\newcommand{\Sierra}{\mathbf{M}_\mathcal{S}}

\usepackage{mathtools}

\newcommand{\minimize}{\operatornamewithlimits{minimize}}

\newcommand{\argmin}{\operatornamewithlimits{arg\,min}}

\newcommand{\R}{\mathbb{R}}

\DeclareMathOperator{\Normal}{\mathcal{N}}

\newcommand{\mat}[1]{\vect{#1}}
\renewcommand{\vec}[1]{\vect{#1}}

\algtext*{EndLoop}% Remove "end _X_" text
\algtext*{EndFor}
\algtext*{EndIf}
\algtext*{EndFunction}

\DeclarePairedDelimiter{\round}\lfloor\rceil

\usepackage{longtable,tabularx,booktabs}
\usepackage[flushleft]{threeparttable}

\makeatletter
\DeclareRobustCommand{\cev}[1]{%
  {\mathpalette\do@cev{#1}}%
}
\newcommand{\do@cev}[2]{%
  \vbox{\offinterlineskip
    \sbox\z@{$\m@th#1 x$}%
    \ialign{##\cr
      \hidewidth\reflectbox{$\m@th#1\vec{}\mkern4mu$}\hidewidth\cr
      \noalign{\kern-\ht\z@}
      $\m@th#1#2$\cr
    }%
  }
}%late
\makeatother

\usepackage{multirow}

\usetikzlibrary{calc}
\usetikzlibrary{shapes.geometric}
\usetikzlibrary{external}
\usetikzlibrary{patterns}
\usetikzlibrary{shapes,arrows,fit}
\usetikzlibrary{positioning}
\usetikzlibrary{arrows.meta, calc, shapes}
\usetikzlibrary{graphs}
\usetikzlibrary{decorations.pathmorphing}
\usetikzlibrary{decorations.pathreplacing}

\usepackage[caption=false]{subfig}

\usepackage{pgfplots}
\pgfplotsset{compat=newest}
\pgfplotsset{every axis legend/.append style={%
cells={anchor=west}}
}
\usetikzlibrary{arrows}
\tikzset{>=stealth'}

\usepgfplotslibrary{fillbetween}
\usepgfplotslibrary{groupplots}

\newcommand{\smallcaps}[1]{\textsc{#1}}

\begin{document}

\title{Cross-Entropy Method Variants for Optimization
\thanks{Code available at \href{https://github.com/mossr/CrossEntropyVariants.jl}{https://github.com/mossr/CrossEntropyVariants.jl}}}

\author{\IEEEauthorblockN{Robert J. Moss}
Stanford University, Computer Science\\
Stanford, CA, 94305\\
mossr@cs.stanford.edu}

\maketitle

\begin{abstract}
The cross-entropy (CE) method is a popular stochastic method for optimization due to its simplicity and effectiveness.
Designed for rare-event simulations where the probability of a target event occurring is relatively small,
the CE-method relies on enough objective function calls to accurately estimate the optimal parameters of the underlying distribution. 
Certain objective functions may be computationally expensive to evaluate, and the CE-method could potentially get stuck in local minima.
This is compounded with the need to have an initial covariance wide enough to cover the design space of interest.
We introduce novel variants of the CE-method to address these concerns.
To mitigate expensive function calls, during optimization we use every sample to build a surrogate model to approximate the objective function.
The surrogate model augments the belief of the objective function with less expensive evaluations.
We use a Gaussian process for our surrogate model to incorporate uncertainty in the predictions which is especially helpful when dealing with sparse data.
To address local minima convergence, we use Gaussian mixture models to encourage exploration of the design space.
We experiment with evaluation scheduling techniques to reallocate true objective function calls earlier in the optimization when the covariance is the largest.
To test our approach, we created a parameterized test objective function with many local minima and a single global minimum. Our test function can be adjusted to control the spread and distinction of the minima.
Experiments were run to stress the cross-entropy method variants and results indicate that the surrogate model-based approach reduces local minima convergence using the same number of function evaluations.
\end{abstract}

\section{Introduction}
The cross-entropy (CE) method is a probabilistic optimization approach that attempts to iteratively fit a distribution to elite samples from an initial input distribution \cite{rubinstein2004cross,rubinstein1999cross}.
The goal is to estimate a rare-event probability by minimizing the \textit{cross-entropy} between the two distributions \cite{de2005tutorial}.
The CE-method has gained popularity in part due to its simplicity in implementation and straightforward derivation.
The technique uses \textit{importance sampling} which introduces a proposal distribution over the rare-events to sample from then re-weights the posterior likelihood by the \textit{likelihood ratio} of the true distribution over the proposal distribution.

There are a few key assumptions that make the CE-method work effectively.
Through random sampling, the CE-method assumes that there are enough objective function evaluations to accurately represent the objective. 
This may not be a problem for simple applications, but can be an issue for computationally expensive objective functions. 
Another assumption is that the initial parameters of the input distribution are wide enough to cover the design space of interest. For the case with a multivariate Gaussian distribution, this corresponds to an appropriate mean and wide covariance.
In rare-event simulations with many local minima, the CE-method can fail to find a global minima especially with sparse objective function evaluations.

This work aims to address the key assumptions of the CE-method.
We introduce variants of the CE-method that use surrogate modeling to approximate the objective function, thus updating the belief of the underlying objective through estimation.
As part of this approach, we introduce evaluation scheduling techniques to reallocate true objective function calls earlier in the optimization when we know the covariance will be large.
The evaluation schedules can be based on a distribution (e.g., the Geometric distribution) or can be prescribed manually depending on the problem.
We also use a Gaussian mixture model representation of the prior distribution as a method to explore competing local optima.
While the use of Gaussian mixture models in the CE-method is not novel, we connect the use of mixture models and surrogate modeling in the CE-method.
This connection uses each elite sample as the mean of a component distribution in the mixture, optimized through a subroutine call to the standard CE-method using the learned surrogate model.
To test our approach, we introduce a parameterized test objective function called \textit{sierra}.
The sierra function is built from a multivariate Gaussian mixture model with many local minima and a single global minimum.
Parameters for the sierra function allow control over both the spread and distinction of the minima.
Lastly, we provide an analysis of the weak areas of the CE-method compared to our proposed variants.

\section{Related Work} \label{sec:related_work}
The cross-entropy method is popular in the fields of operations research, machine learning, and optimization \cite{kochenderfer2015decision,Kochenderfer2019}.
The combination of the cross-entropy method, surrogate modeling, and mixture models has been explored in other work \cite{bardenet2010surrogating}. 
The work in \cite{bardenet2010surrogating} proposed an adaptive grid approach to accelerate Gaussian-process-based surrogate modeling using mixture models as the prior in the cross-entropy method. They showed that a mixture model performs better than a single Gaussian when the objective function is multimodal.
Our work differs in that we augment the ``elite'' samples both by an approximate surrogate model and by a subroutine call to the CE-method using the learned surrogate model.
Other related work use Gaussian processes and a modified cross-entropy method for receding-horizon trajectory optimization \cite{tan2018gaussian}.
Their cross-entropy method variant also incorporates the notion of exploration in the context of path finding applications.
An approach based on \textit{relative entropy}, described in \cref{sec:background_ce}, proposed a model-based stochastic search that seeks to minimize the relative entropy \cite{NIPS2015_5672}. They also explore the use of a simple quadratic surrogate model to approximate the objective function.
Prior work that relate cross-entropy-based adaptive importance sampling with Gaussian mixture models show that a mixture model require less objective function calls than a na\"ive Monte Carlo or standard unimodal cross-entropy-based importance sampling method \cite{kurtz2013cross,wang2016cross}.

\section{Background} \label{sec:background}
This section provides necessary background on techniques used in this work. We provide introductions to cross-entropy and the cross-entropy method, surrogate modeling using Gaussian processes, and multivariate Gaussian mixture models.

\subsection{Cross-Entropy} \label{sec:background_ce}
Before understanding the cross-entropy method, we first must understand the notion of \textit{cross-entropy}.
Cross-entropy is a metric used to measure the distance between two probability distributions, where the distance may not be symmetric \cite{de2005tutorial}.
The distance used to define cross-entropy is called the \textit{Kullback-Leibler (KL) distance} or \textit{KL divergence}.
The KL distance is also called the \textit{relative entropy}, and we can use this to derive the cross-entropy.
Formally, for a random variable $\mat{X} = (X_1, \ldots, X_n)$ with a support of $\mathcal{X}$, the KL distance between two continuous probability density functions $f$ and $g$ is defined to be:
\begin{align*}
    \mathcal{D}(f, g) &= \E_f\left[\log \frac{f(\vec{X})}{g(\vec{X})} \right]\\
                      &= \int\limits_{\vec{x} \in \mathcal{X}} f(\vec{x}) \log f(\vec{x}) d\vec{x} - \int\limits_{\vec{x} \in \mathcal{X}} f(\vec{x}) \log g(\vec{x}) d\vec{x}
\end{align*}
We denote the expectation of some function with respect to a distribution $f$ as $\E_f$.
Minimizing the KL distance $\mathcal{D}$ between our true distribution $f$ and our proposal distribution $g$ parameterized by $\vec{\theta}$, is equivalent to choosing $\vec\theta$ that minimizes the following, called the \textit{cross-entropy}:
\begin{align*}
    H(f,g) &= H(f) + \mathcal{D}(f,g)\\
           &= -\E_f[\log g(\vec{X})] \tag{using KL distance}\\
           &= - \int\limits_{\vec{x} \in \mathcal{X}} f(\vec{x}) \log g(\vec{x} \mid \vec{\theta}) d\vec{x}
\end{align*}
where $H(f)$ denotes the entropy of the distribution $f$ (where we conflate entropy and continuous entropy for convenience).
This assumes that $f$ and $g$ share the support $\mathcal{X}$ and are continuous with respect to $\vec{x}$.
The minimization problem then becomes:
\begin{equation} \label{eq:min}
\begin{aligned}
    \minimize_{\vec{\theta}} & & - \int\limits_{\vec{x} \in \mathcal{X}} f(\vec{x}) \log g(\vec{x} \mid \vec{\theta}) d\vec{x}
\end{aligned}
\end{equation}
Efficiently finding this minimum is the goal of the cross-entropy method algorithm.

\subsection{Cross-Entropy Method} \label{sec:background_cem}
Using the definition of cross-entropy, intuitively the \textit{cross-entropy method} (CEM or CE-method) aims to minimize the cross-entropy between the unknown true distribution $f$ and a proposal distribution $g$ parameterized by $\vec\theta$.
This technique reformulates the minimization problem as a probability estimation problem, and uses adaptive importance sampling to estimate the unknown expectation \cite{de2005tutorial}.
The cross-entropy method has been applied in the context of both discrete and continuous optimization problems \cite{rubinstein1999cross,kroese2006cross}.

The initial goal is to estimate the probability 
\begin{align*}
    \ell = P_{\vec{\theta}}(S(\vec{x}) \ge \gamma)
\end{align*}
where $S$ can the thought of as an objective function of $\vec{x}$, and $\vec{x}$ follows a distribution defined by $g(\vec{x} \mid \vec{\theta})$.
We want to find events where our objective function $S$ is above some threshold $\gamma$.
We can express this unknown probability as the expectation
\begin{align} \label{eq:expect}
    \ell = \E_{\vec{\theta}}[\mathbbm{1}_{(S(\vec{x}) \ge \gamma)}]
\end{align}
where $\mathbbm{1}$ denotes the indicator function.
A straightforward way to estimate \cref{eq:expect} can be done through Monte Carlo sampling.
But for rare-event simulations where the probability of a target event occurring is relatively small, this estimate becomes inadequate.
The challenge of the minimization in \cref{eq:min} then becomes choosing the density function for the true distribution $f(\vec{x})$. 
Importance sampling tells us that the optimal importance sampling density can be reduced to
\begin{align*}
    f^*(\vec{x}) = \frac{\mathbbm{1}_{(S(\vec{x}) \ge \gamma)}g(\vec{x} \mid \vec{\theta})}{\ell}
\end{align*}
thus resulting in the optimization problem:
\begin{align*}
    \vec{\theta}_g^* &= \argmin_{\vec{\theta}_g} - \int\limits_{\vec{x} \in \mathcal{X}} f^*(\vec{x})\log g(\vec{x} \mid \vec{\theta}_g) d\vec{x}\\
                   &= \argmin_{\vec{\theta}_g} - \int\limits_{\vec{x} \in \mathcal{X}} \frac{\mathbbm{1}_{(S(\vec{x}) \ge \gamma)}g(\vec{x} \mid \vec{\theta})}{\ell}\log g(\vec{x} \mid \vec{\theta}_g) d\vec{x}
\end{align*}
Note that since we assume $f$ and $g$ belong to the same family of distributions, we get that $f(\vec{x}) = g(\vec{x} \mid \vec{\theta}_g)$.
Now notice that $\ell$ is independent of $\vec{\theta}_g$, thus we can drop $\ell$ and get the final optimization problem of:
\begin{align} \label{eq:opt}
    \vec{\theta}_g^* &= \argmin_{\vec{\theta}_g} - \int\limits_{\vec{x} \in \mathcal{X}} \mathbbm{1}_{(S(\vec{x}) \ge \gamma)}g(\vec{x} \mid \vec{\theta}) \log g(\vec{x} \mid \vec{\theta}_g) d\vec{x}\\\nonumber
                   &= \argmin_{\vec{\theta}_g} - \E_{\vec{\theta}}[ \mathbbm{1}_{(S(\vec{x}) \ge \gamma)} \log g(\vec{x} \mid \vec{\theta}_g)]
\end{align}

The CE-method uses a multi-level algorithm to estimate $\vec{\theta}_g^*$ iteratively.
The parameter $\vec{\theta}_k$ at iteration $k$ is used to find new parameters $\vec{\theta}_{k^\prime}$ at the next iteration $k^\prime$.
The threshold $\gamma_k$ becomes smaller that its initial value, thus artificially making events \textit{less rare} under $\vec{X} \sim g(\vec{x} \mid \vec{\theta}_k)$.

In practice, the CE-method algorithm requires the user to specify a number of \textit{elite} samples $m_\text{elite}$ which are used when fitting the new parameters for iteration $k^\prime$.
Conveniently, if our distribution $g$ belongs to the \textit{natural exponential family} then the optimal parameters can be found analytically \cite{Kochenderfer2019}. For a multivariate Gaussian distribution parameterized by $\vec{\mu}$ and $\mat{\Sigma}$, the optimal parameters for the next iteration $k^\prime$ correspond to the maximum likelihood estimate (MLE):
\begin{align*}
    \vec{\mu}_{k^\prime} &= \frac{1}{m_\text{elite}} \sum_{i=1}^{m_\text{elite}} \vec{x}_i\\
    \vec{\Sigma}_{k^\prime} &= \frac{1}{m_\text{elite}} \sum_{i=1}^{m_\text{elite}} (\vec{x}_i - \vec{\mu}_{k^\prime})(\vec{x}_i - \vec{\mu}_{k^\prime})^\top
\end{align*}

The cross-entropy method algorithm is shown in \cref{alg:cem}.
For an objective function $S$ and input distribution $g$, the CE-method algorithm will run for $k_\text{max}$ iterations.
At each iteration, $m$ inputs are sampled from $g$ and evaluated using the objective function $S$.
The sampled inputs are denoted by $\mat{X}$ and the evaluated values are denoted by $\mat{Y}$.
Next, the top $m_\text{elite}$ samples are stored in the elite set $\e$, and the distribution $g$ is fit to the elites.
This process is repeated for $k_\text{max}$ iterations and the resulting parameters $\vec{\theta}_{k_\text{max}}$ are returned.
Note that a variety of input distributions for $g$ are supported, but we focus on the multivariate Gaussian distribution and the Gaussian mixture model in this work.

%% CE-method
\begin{algorithm}[ht]
  \begin{algorithmic}
  \Function{CrossEntropyMethod}{}($S, g, m, m_\text{elite}, k_\text{max}$)
    \For {$k \in [1,\ldots,k_\text{max}]$}
        \State $\mat{X} \sim g(\;\cdot \mid \vec{\theta}_k)$ where $\mat{X} \in \R^m$
        \State $\mat{Y} \leftarrow S(\vec{x})$ for $\vec{x} \in \mat{X}$
        \State $\e \leftarrow$ store top $m_\text{elite}$ from $\mat{Y}$
        \State $\vec{\theta}_{k^\prime} \leftarrow \textproc{Fit}(g(\;\cdot \mid \vec{\theta}_k), \e)$
    \EndFor
    \State \Return $g(\;\cdot \mid \vec{\theta}_{k_\text{max}})$
  \EndFunction
  \end{algorithmic}
  \caption{\label{alg:cem} Cross-entropy method.}
\end{algorithm}

\subsection{Mixture Models}
A standard Gaussian distribution is \textit{unimodal} and can have trouble generalizing over data that is \textit{multimodal}.
A \textit{mixture model} is a weighted mixture of component distributions used to represent continuous multimodal distributions \cite{kochenderfer2015decision}.
Formally, a Gaussian mixture model (GMM) is defined by its parameters $\vec{\mu}$ and $\mat{\Sigma}$ and associated weights $\w$ where $\sum_{i=1}^n w_i = 1$. We denote that a random variable $\mat{X}$ is distributed according to a mixture model as $\mat{X} \sim \operatorname{Mixture}(\vec{\mu}, \vec{\Sigma}, \vec{w})$.
The probability density of the GMM then becomes:
%% Mixture model PDF
\begin{gather*}
    P( \mat{X} = \vec{x} \mid \vec{\mu}, \mat{\Sigma}, \vec{w}) = \sum_{i=1}^n w_i \Normal(\vec{x} \mid \vec{\mu}_i, \mat{\Sigma}_i)
\end{gather*}

To fit the parameters of a Gaussian mixture model, it is well known that the \textit{expectation-maximization (EM)} algorithm can be used \cite{dempster1977maximum,aitkin1980mixture}. 
The EM algorithm seeks to find the maximum likelihood estimate of the hidden variable $H$ using the observed data defined by $E$.
Intuitively, the algorithm alternates between an expectation step (E-step) and a maximization step (M-step) to guarantee convergence to a local minima.
A simplified EM algorithm is provide in \cref{alg:em} for reference and we refer to \cite{dempster1977maximum,aitkin1980mixture} for further reading.

%% Expectation Maximization
\begin{algorithm}[ht]
  \begin{algorithmic}
  \Function{ExpectationMaximization}{$H, E, \vec{\theta}$}
    \For{\textbf{E-step}}
        \State Compute $Q(h) = P(H=h \mid E=e, \vec{\theta})$ for each $h$ % (use any probabilistic inference algorithm)
        \State Create weighted points: $(h,e)$ with weight $Q(h)$
    \EndFor
    \For{\textbf{M-step}}
        \State Compute $\mathbf{\hat{\vec{\theta}}}_{\text{MLE}}$
    \EndFor
    \State Repeat until convergence.
    \State \Return $\mathbf{\hat{\vec{\theta}}}_{\text{MLE}}$
  \EndFunction
  \end{algorithmic}
  \caption{\label{alg:em} Expectation-maximization.}
\end{algorithm}

\subsection{Surrogate Models}
In the context of optimization, a surrogate model $\hat{S}$ is used to estimate the true objective function and provide less expensive evaluations.
Surrogate models are a popular approach and have been used to evaluate rare-event probabilities in computationally expensive systems \cite{li2010evaluation,li2011efficient}.
The simplest example of a surrogate model is linear regression.
In this work, we focus on the \textit{Gaussian process} surrogate model.
A Gaussian process (GP) is a distribution over functions that predicts the underlying objective function $S$ and captures the uncertainty of the prediction using a probability distribution \cite{Kochenderfer2019}.
This means a GP can be sampled to generate random functions, which can then be fit to our given data $\mat{X}$.
A Gaussian process is parameterized by a mean function $\m(\mat{X})$ and kernel function $\mat{K}(\mat{X},\mat{X})$, which captures the relationship between data points as covariance values.
We denote a Gaussian process that produces estimates $\hat{\vec{y}}$ as:
\begin{align*}
\hat{\vec{y}} &\sim\mathcal{N}\left(\vec{m}(\mat{X}),\vec{K}(\mat{X},\mat{X})\right)\\
        &= \begin{bmatrix} % Changed `m` to `n`
            \hat{S}(\vec{x}_1), \ldots, \hat{S}(\vec{x}_n)
        \end{bmatrix}
\end{align*}
where
\begin{gather*}
\vec{m}(\mat{X}) = \begin{bmatrix} m(\vec{x}_1), \ldots, m(\vec{x}_n) \end{bmatrix}\\
\vec{K}(\mat{X}, \mat{X}) = \begin{bmatrix}
         k(\vec{x}_1, \vec{x}_1) & \cdots & k(\vec{x}_1, \vec{x}_n)\\
         \vdots & \ddots & \vdots\\
         k(\vec{x}_n, \vec{x}_1) & \cdots & k(\vec{x}_n, \vec{x}_n)
     \end{bmatrix}
\end{gather*}
We use the commonly used zero-mean function $m(\vec{x}_i) = \vec{0}$.
For the kernel function $k(\vec{x}_i, \vec{x}_i)$, we use the squared exponential kernel with variance $\sigma^2$ and characteristic scale-length $\ell$, where larger $\ell$ values increase the correlation between successive data points, thus smoothing out the generated functions. The squared exponential kernel is defined as:
% Isotropic Squared Exponential kernel (covariance): \exp(-\frac{r^2}{2\ell^2})
\begin{align*}
k(\vec{x},\vec{x}^\prime) = \sigma^2\exp\left(- \frac{(\vec{x} - \vec{x}^\prime)^\top(\vec{x} - \vec{x}^\prime)}{2\ell^2}\right)
\end{align*}
We refer to \cite{Kochenderfer2019} for a detailed overview of Gaussian processes and different kernel functions.

\section{Algorithms} \label{sec:algorithms}
We can now describe the cross-entropy method variants introduced in this work.
This section will first cover the main algorithm introduced, the cross-entropy surrogate method (CE-surrogate).
Then we introduce a modification to the CE-surrogate method, namely the cross-entropy mixture method (CE-mixture).
Lastly, we describe various evaluation schedules for redistributing objective function calls over the iterations.

\subsection{Cross-Entropy Surrogate Method} \label{sec:alg_ce_surrogate}
The main CE-method variant we introduce is the cross-entropy surrogate method (CE-surrogate).
The CE-surrogate method is a superset of the CE-method, where the differences lie in the evaluation scheduling and modeling of the elite set using a surrogate model.
The goal of the CE-surrogate algorithm is to address the shortcomings of the CE-method when the number of objective function calls is sparse and the underlying objective function $S$ has multiple local minima.

The CE-surrogate algorithm is shown in \cref{alg:ce_surrogate}.
It takes as input the objective function $S$, the distribution $\M$ parameterized by $\vec{\theta}$, the number of samples $m$, the number of elite samples $m_\text{elite}$, and the maximum iterations $k_\text{max}$.
For each iteration $k$, the number of samples $m$ are redistributed through a call to \smallcaps{EvaluationSchedule}, where $m$ controls the number of true objective function evaluations of $S$. % EvaluationSchedule.
Then, the algorithm samples from $\M$ parameterized by the current $\vec{\theta}_k$ given the adjusted number of samples $m$. % and clamped $m_\text{elite}$.
For each sample in $\mat{X}$, the objective function $S$ is evaluated and the results are stored in $\mat{Y}$.
The top $m_\text{elite}$ evaluations from $\mat{Y}$ are stored in $\e$. 
Using all of the current function evaluations $\mat{Y}$ from sampled inputs $\mat{X}$, a modeled elite set $\bfE$ is created to augment the sparse information provided by a low number of true objective function evaluations.
Finally, the distribution $\M$ is fit to the elite set $\bfE$ and the distribution with the final parameters $\vec{\theta}_{k_\text{max}}$ is returned.

%% CE-surrogate
\begin{algorithm}[ht]
  \begin{algorithmic}
  \Function{CE-Surrogate}{$S$, $\M$, $m$, $m_\text{elite}$, $k_\text{max}$}
    \For {$k \in [1,\ldots,k_\text{max}]$}
        \State $m, m_\text{elite} \leftarrow \textproc{EvaluationSchedule}(k, k_\text{max})$
        \State $\mat{X} \sim \M(\;\cdot \mid \vec{\theta}_k)$ where $\mat{X} \in \R^m$
        \State $\mat{Y} \leftarrow S(\vec{x})$ for $\vec{x} \in \mat{X}$
        \State $\e \leftarrow$ store top $m_\text{elite}$ from $\mat{Y}$
        \State $\bfE \leftarrow \textproc{ModelEliteSet}(\mat{X}, \mat{Y}, \M, \e, m, m_\text{elite})$
        \State $\vec{\theta}_{k^\prime} \leftarrow \textproc{Fit}(\M(\;\cdot \mid \vec{\theta}_k), \bfE)$
    \EndFor
    \State \Return $\M(\;\cdot \mid \vec{\theta}_{k_\text{max}})$
  \EndFunction
  \end{algorithmic}
  \caption{\label{alg:ce_surrogate} Cross-entropy surrogate method.}
\end{algorithm}

The main difference between the standard CE-method and the CE-surrogate variant lies in the call to \smallcaps{ModelEliteSet}.
The motivation is to use \textit{all} of the already evaluated objective function values $\mat{Y}$ from a set of sampled inputs $\mat{X}$.
This way the expensive function evaluations---otherwise discarded---can be used to build a surrogate model of the underlying objective function.
First, a surrogate model $\surrogate$ is constructed from the samples $\mat{X}$ and true objective function values $\mat{Y}$.
We used a Gaussian process with a specified kernel and optimizer, but other surrogate modeling techniques such as regression with basis functions can be used.
We chose a Gaussian process because it incorporates probabilistic uncertainty in the predictions, which may more accurately represent our objective function, or at least be sensitive to over-fitting to sparse data.
Now we have an approximated objective function $\surrogate$ that we can inexpensively call. 
We sample $10m$ values from the distribution $\M$ and evaluate them using the surrogate model.
We then store the top $10m_\text{elite}$ values from the estimates $\mathbf{\hat{\mat{Y}}}_\text{m}$.
We call these estimated elite values $\e_\text{model}$ the \textit{model-elites}.
The surrogate model is then passed to \smallcaps{SubEliteSet}, which returns more estimates for elite values.
Finally, the elite set $\bfE$ is built from the true-elites $\e$, the model-elites $\e_\text{model}$, and the subcomponent-elites $\e_\text{sub}$.
The resulting concatenated elite set $\bfE$ is returned.

\begin{algorithm}[ht]
  \begin{algorithmic}
  \Function{ModelEliteSet}{$\mat{X}, \mat{Y}, \M, \e, m, m_\text{elite}$}
    % Fit to entire population!
    \State $\surrogate \leftarrow \textproc{GaussianProcess}(\mat{X}, \mat{Y}, \text{kernel}, \text{optimizer})$ % Squared exponential, NelderMead
    \State $\mat{X}_\text{m} \sim \M(\;\cdot \mid \vec{\theta}_k)$ where $\mat{X}_\text{m} \in \R^{10m}$
    \State $\mathbf{\hat{\mat{Y}}}_\text{m} \leftarrow \surrogate(\vec{x}_\text{m})$ for $\vec{x}_\text{m} \in \mat{X}_\text{m}$
    \State $\e_\text{model} \leftarrow$ store top $10m_\text{elite}$ from $\mathbf{\hat{\mat{Y}}}_\text{m}$
    \State $\e_\text{sub} \leftarrow \textproc{SubEliteSet}(\surrogate, \M, \e)$
    \State $\bfE \leftarrow \{ \e \} \cup \{ \e_\text{model} \} \cup \{ \e_\text{sub} \}$ \algorithmiccomment{elite set}
    \State \Return $\bfE$
  \EndFunction
  \end{algorithmic}
  \caption{\label{alg:model_elite_set} Modeling elite set.}
\end{algorithm}

To encourage exploration of promising areas of the design space, the algorithm \smallcaps{SubEliteSet} focuses on the already marked true-elites $\e$.
Each elite $e_x \in \e$ is used as the mean of a new multivariate Gaussian distribution with covariance inherited from the distribution $\M$.
The collection of subcomponent distributions is stored in $\m$.
The idea is to use the information given to us by the true-elites to emphasize areas of the design space that look promising.
For each distribution $\m_i \in \m$ we run a subroutine call to the standard CE-method to fit the distribution $\m_i$ using the surrogate model $\surrogate$. 
Then the best objective function value is added to the subcomponent-elite set $\e_\text{sub}$, and after iterating the full set is returned.
Note that we use $\theta_\text{CE}$ to denote the parameters for the CE-method algorithm.
In our case, we recommend using a small $k_\text{max}$ of around $2$ so the subcomponent-elites do not over-fit to the surrogate model but have enough CE-method iterations to tend towards optimal.

\begin{algorithm}[ht]
  \begin{algorithmic}
  \Function{SubEliteSet}{$\surrogate, \M, \e$}
    \State $\e_\text{sub} \leftarrow \emptyset$
    \State $\m \leftarrow \{ e_x \in \e \mid \Normal(e_x, \M.\Sigma) \}$
    \For {$\m_i \in \m$}
        \State $\m_i \leftarrow \textproc{CrossEntropyMethod}(\surrogate, \m_i \mid \theta_{\text{CE}})$
        \State $\e_\text{sub} \leftarrow \{\e_\text{sub}\} \cup \{\textproc{Best}(\m_i)\}$
    \EndFor
    \State \Return $\e_\text{sub}$
  \EndFunction
  \end{algorithmic}
  \caption{\label{alg:sub_elite_set} Subcomponent elite set.}
\end{algorithm}

\subsection{Cross-Entropy Mixture Method} \label{sec:alg_ce_mixture}
We refer to the variant of our CE-surrogate method that takes an input \textit{mixture model} $\M$ as the cross-entropy mixture method (CE-mixture).
The CE-mixture algorithm is identical to the CE-surrogate algorithm, but calls a custom \smallcaps{Fit} function to fit a mixture model to the elite set $\bfE$.
The input distribution $\M$ is cast to a mixture model using the subcomponent distributions $\m$ as the components of the mixture.
We use the default uniform weighting for each mixture component.
The mixture model $\M$ is then fit using the expectation-maximization algorithm shown in \cref{alg:em}, and the resulting distribution is returned.
The idea is to use the distributions in $\m$ that are centered around each true-elite as the components of the casted mixture model.
Therefore, we would expect better performance of the CE-mixture method when the objective function has many competing local minima.
Results in \cref{sec:results} aim to show this behavior.

%% CE-mixture (fit)
\begin{algorithm}[ht]
  \begin{algorithmic}
  \Function{Fit}{$\M, \m, \bfE$}
    \State $\M \leftarrow \operatorname{Mixture}( \m )$
    \State $\mathbf{\hat{\vec{\theta}}} \leftarrow \textproc{ExpectationMaximization}(\M, \bfE)$
    \State \Return $\M(\;\cdot \mid \mathbf{\hat{\vec{\theta}}})$
  \EndFunction
  \end{algorithmic}
  \caption{\label{alg:ce_mixture_fit} Fitting mixture models (used by CE-mixture).}
\end{algorithm}

\subsection{Evaluation Scheduling} \label{sec:alg_eval_schedule}
Given the nature of the CE-method, we expect the covariance to shrink over time, thus resulting in a solution with higher confidence.
Yet if each iteration is given the same number of objective function evaluations $m$, there is the potential for elite samples from early iterations dominating the convergence.
Therefore, we would like to redistribute the objective function evaluations throughout the iterations to use more truth information early in the process.
We call these heuristics \textit{evaluation schedules}.
One way to achieve this is to reallocate the evaluations according to a Geometric distribution.
Evaluation schedules can also be ad-hoc and manually prescribed based on the current iteration.

We provide the evaluation schedule we use that follows a Geometric distribution with parameter $p$ in \cref{alg:evaluation_schedule}.
We denote $G \sim \Geo(p)$ to be a random variable that follows a truncated Geometric distribution with the probability mass function $p_G(k) = p(1 - p)^k$ for $k \in \{0, 1, 2, \ldots, k_\text{max}\}$. % Geo(p) PMF
Note the use of the integer rounding function (e.g., $\round{x}$), which we later have to compensate for towards the final iterations.
Results in \cref{sec:results} compare values of $p$ that control the redistribution of evaluations.

%% EvaluationSchedule
\begin{algorithm}[ht]
  \begin{algorithmic}
  \Function{EvaluationSchedule}{$k, k_\text{max}$}
    \State $G \sim \Geo(p)$
    \State $N_\text{max} \leftarrow k_\text{max} \cdot m$
    \State $m \leftarrow \round{N_\text{max} \cdot p_G(k)}$
    \If{$k = k_\text{max}$}
        \State $s \leftarrow \displaystyle\sum_{i=1}^{k_\text{max}-1} \round{N_\text{max} \cdot p_G(i)}$
        \State $m \leftarrow \min(N_\text{max} - s, N_\text{max} - m)$
    \EndIf
    \State $m_\text{elite} \leftarrow \min(m_\text{elite}, m)$
    \State \Return ($m, m_\text{elite}$) 
  \EndFunction
  \end{algorithmic}
  \caption{\label{alg:evaluation_schedule} Evaluation schedule using a Geometric distr.}
\end{algorithm}

\section{Experiments} \label{sec:experiments}
In this section, we detail the experiments we ran to compare the CE-method variants and evaluation schedules.
We first introduce a test objective function we created to stress the issue of converging to local minima. 
We then describe the experimental setup for each of our experiments and provide an analysis and results.

\subsection{Test Objective Function Generation}
\begin{figure*}[!t]
  \centering
  \resizebox{0.8\textwidth}{!}{\input{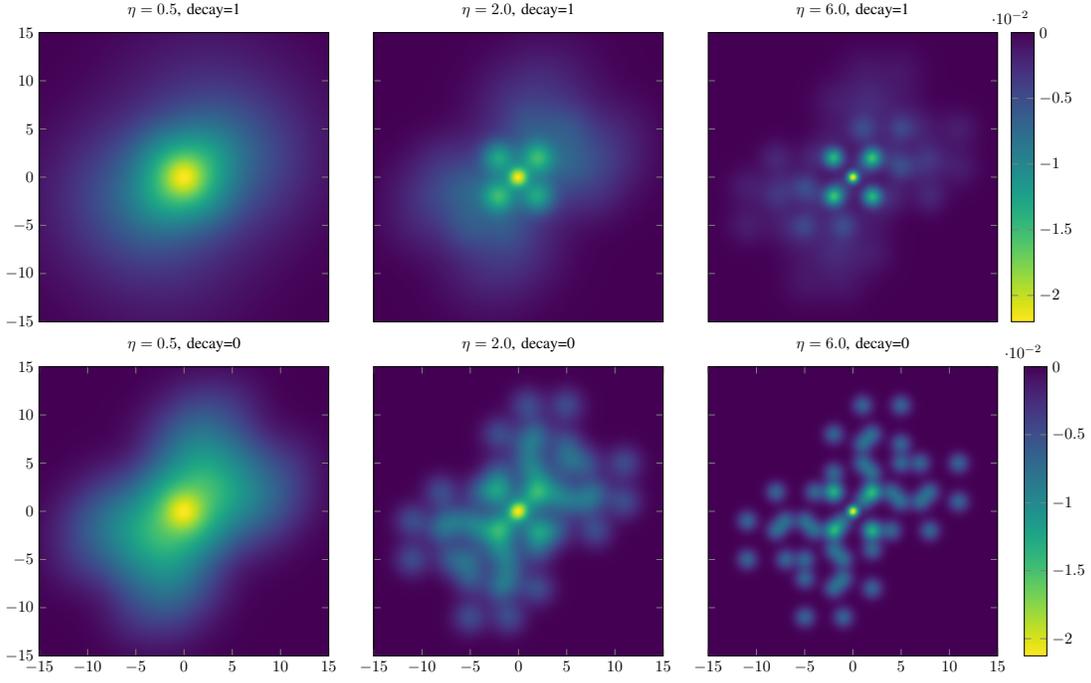}}
  \caption{
    \label{fig:sierra}
    Example test objective functions generated using the sierra function.  
  }
\end{figure*}
To stress the cross-entropy method and its variants, we created a test objective function called \textit{sierra} that is generated from a mixture model comprised of $49$ multivariate Gaussian distributions.
We chose this construction so that we can use the negative peeks of the component distributions as local minima and can force a global minimum centered at our desired $\mathbf{\tilde{\vec{\mu}}}$.
The construction of the sierra test function can be controlled by parameters that define the spread of the local minima.
We first start with the center defined by a mean vector $\mathbf{\tilde{\vec{\mu}}}$ and we use a common covariance $\mathbf{\tilde{\mat{\Sigma}}}$:
\begin{align*}
    \mathbf{\tilde{\vec{\mu}}} &= [\mu_1, \mu_2], \quad \mathbf{\tilde{\mat{\Sigma}}} = \begin{bmatrix}\sigma & 0\\ 0 & \sigma \end{bmatrix}
\end{align*}
Next, we use the parameter $\delta$ that controls the clustered distance between symmetric points:
\begin{align*}
    \mat{G} &= \left\{[+\delta, +\delta], [+\delta, -\delta], [-\delta, +\delta], [-\delta, -\delta]\right\}
\end{align*}
We chose points $\mat{P}$ to fan out the clustered minima relative to the center defined by $\mathbf{\tilde{\vec{\mu}}}$:
\begin{align*}
    \mat{P} &= \left\{[0, 0], [1, 1], [2, 0], [3, 1], [0, 2], [1, 3]\right\}
\end{align*}
The vector $\vec{s}$ is used to control the $\pm$ distance to create an `s' shape comprised of minima, using the standard deviation $\sigma$:
$\vec{s} = \begin{bmatrix}+\sigma, -\sigma \end{bmatrix}$.
We set the following default parameters: standard deviation $\sigma=3$, spread rate $\eta=6$, and cluster distance $\delta=2$.
We can also control if the local minima clusters ``decay'', thus making those local minima less distinct (where $\text{decay} \in \{0, 1\})$.
The parameters that define the sierra function are collected into $\vec{\theta} = \langle \mathbf{\tilde{\vec{\mu}}}, \mathbf{\tilde{\mat{\Sigma}}}, \mat{G}, \mat{P}, \vec{s} \rangle$.
Using these parameters, we can define the mixture model used by the sierra function as:
\begin{gather*}
    \Sierra \sim \operatorname{Mixture}\left(\left\{ \vec{\theta} ~\Big|~ \Normal\left(\vec{g} +  s\vec{p}_i + \mathbf{\tilde{\vec{\mu}}},\; \mathbf{\tilde{\mat{\Sigma}}} \cdot i^{\text{decay}}/\eta \right) \right\} \right)\\
    \text{for } (\vec{g}, \vec{p}_i, s) \in (\mat{G}, \mat{P}, \vec{s})
\end{gather*}
We add a final component to be our global minimum centered at $\mathbf{\tilde{\vec{\mu}}}$ and with a covariance scaled by $\sigma\eta$. Namely, the global minimum is $\vec{x}^* = \E[\Normal(\mathbf{\tilde{\vec{\mu}}}, \mathbf{\tilde{\mat{\Sigma}}}/(\sigma\eta))] = \mathbf{\tilde{\vec{\mu}}}$.
We can now use this constant mixture model with $49$ components and define the sierra objective function $\mathcal{S}(\vec{x})$ to be the negative probability density of the mixture at input $\vec{x}$ with uniform weights:

\begin{align*}
    \mathcal{S}(\vec{x}) &= -P(\Sierra = \vec{x}) = -\frac{1}{|\Sierra|}\sum_{j=1}^{n}\Normal(\vec{x} \mid \vec{\mu}_j, \mat{\Sigma}_j)
\end{align*}
An example of six different objective functions generated using the sierra function are shown in \cref{fig:sierra}, sweeping over the spread rate $\eta$, with and without decay.

\subsection{Experimental Setup} \label{sec:experiment_setup}
Experiments were run to stress a variety of behaviors of each CE-method variant.
The experiments are split into two categories: algorithmic and scheduling.
The algorithmic category aims to compare features of each CE-method variant while holding common parameters constant (for a better comparison).
While the scheduling category experiments with evaluation scheduling heuristics.

\begin{figure*}[!t]
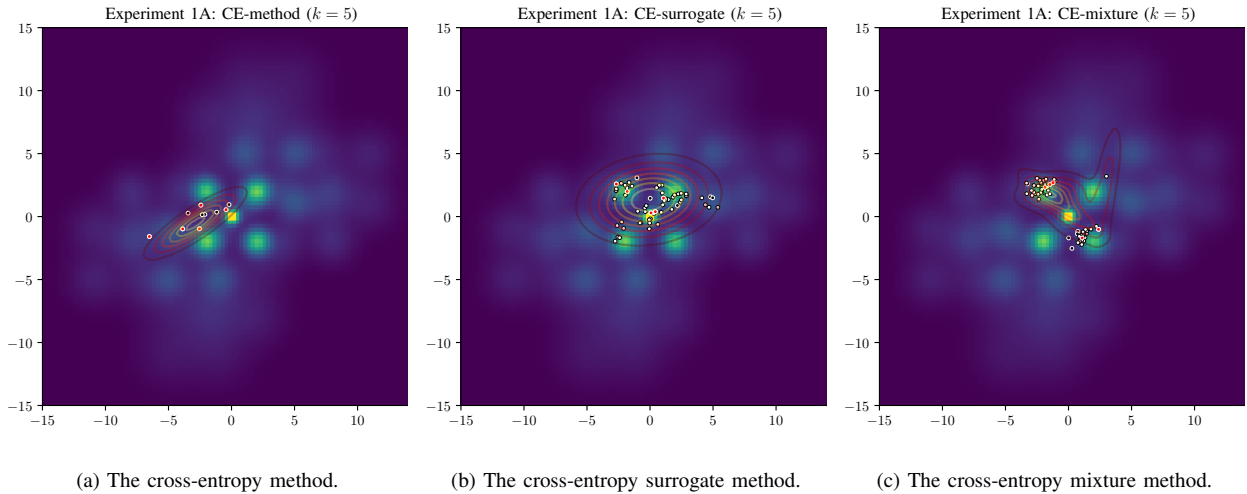

  \centering
    \subfloat[The cross-entropy method.]{%
    \resizebox{0.3\textwidth}{!}{\input{k5_ce_method.pgf}}
    }
    \subfloat[The cross-entropy surrogate method.]{%
    \resizebox{0.3\textwidth}{!}{\input{k5_ce_surrogate.pgf}}
    }
    \subfloat[The cross-entropy mixture method.]{%
    \resizebox{0.3\textwidth}{!}{\input{k5_ce_mixture.pgf}}
    }
  \caption{
    \label{fig:k5} Iteration $k=5$ illustrated for each algorithm. The covariance is shown by the contours.
  } 
\end{figure*}

Because the algorithms are stochastic, we run each experiment with 50 different random number generator seed values.
To evaluate the performance of the algorithms in their respective experiments, we define three metrics.
First, we define the average ``optimal'' value $\bar{b}_v$ to be the average of the best so-far objective function value (termed ``optimal'' in the context of each algorithm). Again, we emphasize that we average over the 50 seed values to gather meaningful statistics.
Another metric we monitor is the average distance to the true global optimal $\bar{b}_d = \norm{\vec{b}_{\vec{x}} - \vec{x}^*}$, where $\vec{b}_{\vec{x}}$ denotes the $\vec{x}$-value associated with the ``optimal''.
We make the distinction between these metrics to show both ``closeness'' in \textit{value} to the global minimum and ``closeness'' in the \textit{design space} to the global minimum.
Our final metric looks at the average runtime of each algorithm, noting that our goal is to off-load computationally expensive objective function calls to the surrogate model.

For all of the experiments, we use a common setting of the following parameters for the sierra test function (shown in the top-right plot in \cref{fig:sierra}):
\begin{equation*}
    (\mathbf{\tilde{\vec{\mu}}} =[0,0],\; \sigma=3,\; \delta=2,\; \eta=6,\; \text{decay} = 1)
\end{equation*}

\subsubsection{Algorithmic Experiments} \label{sec:alg_experiments}
We run three separate algorithmic experiments, each to test a specific feature.
For our first algorithmic experiment (1A), we want to test each algorithm when the user-defined mean is centered at the global minimum and the covariance is arbitrarily wide enough to cover the design space.
Let $\M$ be a distribution parameterized by $\vec{\theta} = (\vec{\mu}, \mat{\Sigma})$, and for experiment (1A) we set the following:
% CE-mixture mean and covariance (1A)
\begin{equation*}
    \vec\mu^{(\text{1A})} = [0, 0] \qquad
    \mat\Sigma^{(\text{1A})} = \begin{bmatrix}
        200 & 0\\
        0 & 200
    \end{bmatrix}
\end{equation*}

For our second algorithmic experiment (1B), we test a mean that is far off-centered with a wider covariance:
% CE-mixture mean and covariance (1B)
\begin{equation*}
    \vec\mu^{(\text{1B})} = [-50, -50] \qquad
    \mat\Sigma^{(\text{1B})} = \begin{bmatrix}
        2000 & 0\\
        0 & 2000
    \end{bmatrix}
\end{equation*}
This experiment is used to test the ``exploration'' of the CE-method variants introduced in this work.
In experiments (1A) and (1B), we set the following common parameters across each CE-method variant:
%% CE-mixture hyperparameter settings
\begin{equation*}
    (k_\text{max} = 10,\; m=10,\; m_\text{elite}=5)^{(\text{1A,1B})}
\end{equation*}
This results in $m\cdot k_\text{max} = 100$ objective function evaluations, which we define to be \textit{relatively} low.

For our third algorithmic experiment (1C), we want to test how each variant responds to an extremely low number of function evaluations.
This sparse experiment sets the common CE-method parameters to:
% CE-method params (1C)
\begin{equation*}
    (k_\text{max} = 10,\; m=5,\; m_\text{elite}=3)^{(\text{1C})}
\end{equation*}
This results in $m\cdot k_\text{max} = 50$ objective function evaluations, which we defined to be \textit{extremely} low.
We use the same mean and covariance defined for experiment (1A):
\begin{equation*}
    \vec\mu^{(\text{1C})} = [0, 0] \qquad
    \mat\Sigma^{(\text{1C})} = \begin{bmatrix}
        200 & 0\\
        0 & 200
    \end{bmatrix}
\end{equation*}

\subsubsection{Scheduling Experiments} \label{sec:schedule_experiments}
In our final experiment (2), we test the evaluation scheduling heuristics which are based on the Geometric distribution.
We sweep over the parameter $p$ that determines the Geometric distribution which controls the redistribution of objective function evaluations.
In this experiment, we compare the CE-surrogate methods using the same setup as experiment (1B), namely the far off-centered mean.
We chose this setup to analyze exploration schemes when given very little information about the true objective function.

\subsection{Results and Analysis} \label{sec:results}

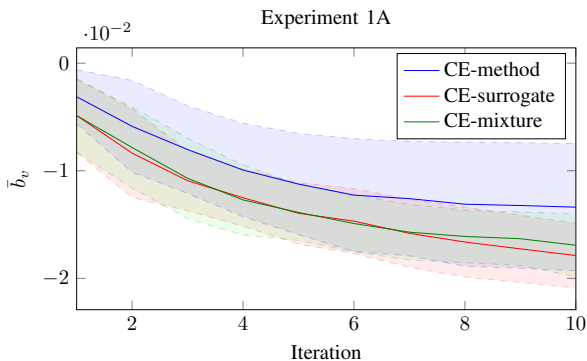
\begin{figure}[!hb]
  % \centering
  \resizebox{0.9\columnwidth}{!}{\begin{tikzpicture}[]
\begin{axis}[height = {6cm}, legend style = {}, ylabel = {$\bar{b}_v$}, title = {Experiment 1A}, xmin = {1}, xmax = {10}, xlabel = {Iteration}, width = {10cm}]\addplot+ [mark = {none}, blue]coordinates {
(1.0, -0.0031201098066532405)
(2.0, -0.005870857044435953)
(3.0, -0.008029959839914747)
(4.0, -0.009931463697053772)
(5.0, -0.011237543181515833)
(6.0, -0.012262911549891264)
(7.0, -0.012592946518249947)
(8.0, -0.013108380477455581)
(9.0, -0.013229171614954224)
(10.0, -0.013384552992011585)
};
\addlegendentry{CE-method}
\addplot+ [mark = {none}, red]coordinates {
(1.0, -0.004867620923558716)
(2.0, -0.008359447446686907)
(3.0, -0.010922475433594345)
(4.0, -0.012512144731967456)
(5.0, -0.013955455243209767)
(6.0, -0.014683807740700328)
(7.0, -0.015815892823378606)
(8.0, -0.016625543566398643)
(9.0, -0.017259993731152726)
(10.0, -0.017879391835753086)
};
\addlegendentry{CE-surrogate}
\addplot+ [mark = {none}, green!50!black]coordinates {
(1.0, -0.004867620923558716)
(2.0, -0.007838400186402038)
(3.0, -0.010702093990244235)
(4.0, -0.012711538858683289)
(5.0, -0.013873504611916936)
(6.0, -0.014905868943295396)
(7.0, -0.015707648145471702)
(8.0, -0.016105669536789855)
(9.0, -0.016316854535311606)
(10.0, -0.016924208538882466)
};
\addlegendentry{CE-mixture}
\addplot+ [mark = {none}, dashed, blue, name path=Aplus, opacity=0.2]coordinates {
(1.0, -0.0006047180244641553)
(2.0, -0.0015997728585300944)
(3.0, -0.003961121055630016)
(4.0, -0.005585741144704704)
(5.0, -0.0065318460375573)
(6.0, -0.007015918213275647)
(7.0, -0.007277530944143609)
(8.0, -0.007347028008143184)
(9.0, -0.007393699423113323)
(10.0, -0.007469733398611503)
};
\addplot+ [mark = {none}, dashed, blue, name path=Aminus, opacity=0.2]coordinates {
(1.0, -0.005635501588842326)
(2.0, -0.010141941230341813)
(3.0, -0.012098798624199478)
(4.0, -0.01427718624940284)
(5.0, -0.015943240325474367)
(6.0, -0.01750990488650688)
(7.0, -0.017908362092356286)
(8.0, -0.018869732946767977)
(9.0, -0.019064643806795126)
(10.0, -0.019299372585411666)
};
\addplot+ [mark = {none}, dashed, red, name path=Bplus, opacity=0.2]coordinates {
(1.0, -0.0014620695871026)
(2.0, -0.004236985725206302)
(3.0, -0.008120707593571749)
(4.0, -0.009862374242251448)
(5.0, -0.01110182392044447)
(6.0, -0.011659906910015172)
(7.0, -0.012664071193490563)
(8.0, -0.013375680889737156)
(9.0, -0.014139327969562021)
(10.0, -0.014872302762209612)
};
\addplot+ [mark = {none}, dashed, red, name path=Bminus, opacity=0.2]coordinates {
(1.0, -0.008273172260014831)
(2.0, -0.012481909168167512)
(3.0, -0.013724243273616942)
(4.0, -0.015161915221683465)
(5.0, -0.016809086565975063)
(6.0, -0.017707708571385483)
(7.0, -0.01896771445326665)
(8.0, -0.01987540624306013)
(9.0, -0.020380659492743432)
(10.0, -0.02088648090929656)
};
\addplot+ [mark = {none}, dashed, green!50!black, name path=Cplus, opacity=0.2]coordinates {
(1.0, -0.0014620695871026)
(2.0, -0.004070861982928433)
(3.0, -0.006957718362033021)
(4.0, -0.009460796991808296)
(5.0, -0.011276196842049358)
(6.0, -0.012279100123204593)
(7.0, -0.013125278973715203)
(8.0, -0.01366560841237955)
(9.0, -0.013839465448164053)
(10.0, -0.014000450783072643)
};
\addplot+ [mark = {none}, dashed, green!50!black, name path=Cminus, opacity=0.2]coordinates {
(1.0, -0.008273172260014831)
(2.0, -0.011605938389875644)
(3.0, -0.01444646961845545)
(4.0, -0.015962280725558282)
(5.0, -0.016470812381784515)
(6.0, -0.017532637763386198)
(7.0, -0.0182900173172282)
(8.0, -0.01854573066120016)
(9.0, -0.01879424362245916)
(10.0, -0.01984796629469229)
};
\addplot[blue!80, fill opacity=0.1] fill between[of=Aplus and Aminus];
                    \addplot[red!80, fill opacity=0.1] fill between[of=Bplus and Bminus];
                    \addplot[green!80, fill opacity=0.1] fill between[of=Cplus and Cminus];
\end{axis}

\end{tikzpicture}}
  \caption{
    \label{fig:experiment_1a}
    Average optimal value for experiment (1A) when the initial mean is centered at the global minimum and the covariance sufficiently covers the design space.
  }
\end{figure}

\Cref{fig:experiment_1a} shows the average value of the current optimal $\bar{b}_v$ for the three algorithms for experiment (1A). 
One standard deviation is plotted in the shaded region.
Notice that the standard CE-method converges to a local minima before $k_\text{max}$ is reached.
Both CE-surrogate method and CE-mixture stay below the standard CE-method curve, highlighting the mitigation of convergence to local minima.
Minor differences can be seen between CE-surrogate and CE-mixture, differing slightly towards the tail in favor of CE-surrogate.
The average runtime of the algorithms along with the performance metrics are shown together for each experiment in \cref{tab:results}.

\begin{table}[!ht]
    \centering
    \caption{\label{tab:results} Experimental results.}
    \begin{tabular}{cllll} % p{3cm}
    \toprule
    \textbf{Exper.} & \textbf{Algorithm} & \textbf{Runtime} & $\bar{b}_v$ & $\bar{b}_d$\\
    \midrule
    \multirow{3}{*}{1A} & CE-method & \textbf{0.029 $\operatorname{s}$} & $-$0.0134 & 23.48\\
    &CE-surrogate & 1.47 $\operatorname{s}$ & \textbf{\boldmath$-$0.0179} & \textbf{12.23}\\
    &CE-mixture & 9.17 $\operatorname{s}$ & $-$0.0169 & 16.87\\
    \midrule
    \multirow{3}{*}{1B} & CE-method & \textbf{0.046 $\operatorname{s}$} & $-$0.0032 & 138.87\\
    &CE-surrogate & 11.82 $\operatorname{s}$ & \textbf{\boldmath$-$0.0156} & \textbf{18.24}\\
    &CE-mixture & 28.10 $\operatorname{s}$ & $-$0.0146 & 33.30\\
    \midrule
    \multirow{3}{*}{1C} & CE-method & \textbf{0.052 $\operatorname{s}$} & $-$0.0065 & 43.14\\
    &CE-surrogate & 0.474 $\operatorname{s}$ & \textbf{\boldmath$-$0.0156} & \textbf{17.23}\\
    &CE-mixture & 2.57 $\operatorname{s}$ & $-$0.0146 & 22.17\\
    \midrule
    \multirow{3}{*}{2} & CE-surrogate, $\operatorname{Uniform}$ & --- & \textbf{\boldmath$-$0.0193} & \textbf{8.53}\\
    &CE-surrogate, $\Geo(0.1)$ & {\color{gray}---} & $-$0.0115 & 25.35\\
    &CE-surrogate, $\Geo(0.2)$ & {\color{gray}---} & $-$0.0099 & 27.59\\
    &CE-surrogate, $\Geo(0.3)$ & {\color{gray}---} & $-$0.0089 & 30.88\\
    \bottomrule
    & & & \multicolumn{2}{l}{$-\text{0.0220} \approx\vec{x}^*$}\\
    \end{tabular}
\end{table}

An apparent benefit of the standard CE-method is in its simplicity and speed.
As shown in \cref{tab:results}, the CE-method is the fastest approach by about 2-3 orders of magnitude compared to CE-surrogate and CE-mixture.
The CE-mixture method is notably the slowest approach.
Although the runtime is also based on the objective function being tested, recall that we are using the same number of true objective function calls in each algorithm, and the metrics we are concerned with in optimization are to minimize $\bar{b}_v$ and $\bar{b}_d$.
We can see that the CE-surrogate method consistently out performs the other methods.
Surprisingly, a uniform evaluation schedule performs the best even in the sparse scenario where the initial mean is far away from the global optimal.

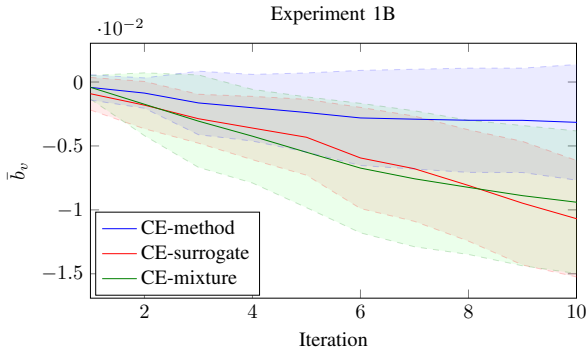
\begin{figure}[!ht]
  % \centering
  \resizebox{0.9\columnwidth}{!}{\begin{tikzpicture}[]
\begin{axis}[height = {6cm}, legend style = {{at={(0.01,0.01)},anchor=south west}}, ylabel = {$\bar{b}_v$}, title = {Experiment 1B}, xmin = {1}, xmax = {10}, xlabel = {Iteration}, width = {10cm}]\addplot+ [mark = {none}, blue]coordinates {
(1.0, -0.00042122778028887684)
(2.0, -0.0008654798596108782)
(3.0, -0.0016348513086736408)
(4.0, -0.002010511770819231)
(5.0, -0.0023846629522817613)
(6.0, -0.002809885878600046)
(7.0, -0.002905905701518952)
(8.0, -0.0029916991837182685)
(9.0, -0.0029962938587671283)
(10.0, -0.003153108797753543)
(11.0, -0.003200678101250048)
(12.0, -0.0032007316164522825)
(13.0, -0.0032177866798048533)
(14.0, -0.0032191390738716737)
(15.0, -0.0032191390738716737)
};
\addlegendentry{CE-method}
\addplot+ [mark = {none}, red]coordinates {
(1.0, -0.0009160031451805364)
(2.0, -0.0018071173179996145)
(3.0, -0.0028698649829150672)
(4.0, -0.0036008686370057694)
(5.0, -0.004320368487047906)
(6.0, -0.0059393295586773285)
(7.0, -0.00678273558549222)
(8.0, -0.008091271531196513)
(9.0, -0.009476259763406172)
(10.0, -0.010690011244082556)
(11.0, -0.01147307819315019)
(12.0, -0.012321804398558118)
(13.0, -0.013183297254110658)
(14.0, -0.01367136706559999)
(15.0, -0.015608547105351183)
};
\addlegendentry{CE-surrogate}
\addplot+ [mark = {none}, green!50!black]coordinates {
(1.0, -0.000412120500565601)
(2.0, -0.001738946222088931)
(3.0, -0.003059216164423346)
(4.0, -0.004236829328663139)
(5.0, -0.00549335742167539)
(6.0, -0.006732282457237139)
(7.0, -0.007568310870316352)
(8.0, -0.008240658426026835)
(9.0, -0.008888142456563063)
(10.0, -0.009401616408731146)
(11.0, -0.009887198873674374)
(12.0, -0.010263372352459069)
(13.0, -0.010744544183763844)
(14.0, -0.01126495335687492)
};
\addlegendentry{CE-mixture}
\addplot+ [mark = {none}, dashed, blue, name path=Aplus, opacity=0.2]coordinates {
(1.0, 0.0005655849280861258)
(2.0, 0.00032175256846091627)
(3.0, 0.0008497895809695827)
(4.0, 0.0005963365316760945)
(5.0, 0.0007097290570611372)
(6.0, 0.0009103115533052007)
(7.0, 0.0009992775069641918)
(8.0, 0.0010856946682476802)
(9.0, 0.0010854983468129714)
(10.0, 0.0013713579241315613)
(11.0, 0.001440724016786289)
(12.0, 0.001440743614448835)
(13.0, 0.0014905872639687503)
(14.0, 0.0014947257311029132)
(15.0, 0.0014947257311029132)
};
\addplot+ [mark = {none}, dashed, blue, name path=Aminus, opacity=0.2]coordinates {
(1.0, -0.0014080404886638795)
(2.0, -0.0020527122876826728)
(3.0, -0.0041194921983168644)
(4.0, -0.004617360073314557)
(5.0, -0.00547905496162466)
(6.0, -0.006530083310505292)
(7.0, -0.006811088910002096)
(8.0, -0.007069093035684217)
(9.0, -0.007078086064347228)
(10.0, -0.007677575519638647)
(11.0, -0.007842080219286385)
(12.0, -0.0078422068473534)
(13.0, -0.007926160623578458)
(14.0, -0.00793300387884626)
(15.0, -0.00793300387884626)
};
\addplot+ [mark = {none}, dashed, red, name path=Bplus, opacity=0.2]coordinates {
(1.0, 0.0003587132390718365)
(2.0, 4.3909388608471536e-5)
(3.0, -0.0009563349640724199)
(4.0, -0.0011219410909393585)
(5.0, -0.0013518441148593534)
(6.0, -0.001982073189738983)
(7.0, -0.0026645191389055127)
(8.0, -0.003734694891290487)
(9.0, -0.004642661716129139)
(10.0, -0.006134757213424801)
(11.0, -0.00703998642994086)
(12.0, -0.007720644278804806)
(13.0, -0.008604485772167086)
(14.0, -0.009176232630158578)
(15.0, -0.011167141396342426)
};
\addplot+ [mark = {none}, dashed, red, name path=Bminus, opacity=0.2]coordinates {
(1.0, -0.0021907195294329092)
(2.0, -0.0036581440246077008)
(3.0, -0.004783395001757715)
(4.0, -0.006079796183072181)
(5.0, -0.007288892859236459)
(6.0, -0.009896585927615675)
(7.0, -0.010900952032078927)
(8.0, -0.01244784817110254)
(9.0, -0.014309857810683207)
(10.0, -0.015245265274740311)
(11.0, -0.01590616995635952)
(12.0, -0.016922964518311427)
(13.0, -0.01776210873605423)
(14.0, -0.0181665015010414)
(15.0, -0.02004995281435994)
};
\addplot+ [mark = {none}, dashed, green!50!black, name path=Cplus, opacity=0.2]coordinates {
(1.0, 0.000514959877286993)
(2.0, 0.0007179203270678671)
(3.0, 0.0005549019657232362)
(4.0, -0.0005752709235030138)
(5.0, -0.0011618180963984875)
(6.0, -0.001668216676329902)
(7.0, -0.0022504153689073573)
(8.0, -0.002992159232448518)
(9.0, -0.0034027054262190494)
(10.0, -0.003824399895506824)
(11.0, -0.004146653236500213)
(12.0, -0.00467764930747694)
(13.0, -0.0051020269786564935)
(14.0, -0.005600974474109703)
};
\addplot+ [mark = {none}, dashed, green!50!black, name path=Cminus, opacity=0.2]coordinates {
(1.0, -0.001339200878418195)
(2.0, -0.004195812771245729)
(3.0, -0.006673334294569929)
(4.0, -0.007898387733823266)
(5.0, -0.009824896746952291)
(6.0, -0.011796348238144376)
(7.0, -0.012886206371725346)
(8.0, -0.013489157619605152)
(9.0, -0.014373579486907078)
(10.0, -0.014978832921955468)
(11.0, -0.015627744510848536)
(12.0, -0.0158490953974412)
(13.0, -0.016387061388871194)
(14.0, -0.016928932239640135)
};
\addplot[blue!80, fill opacity=0.1] fill between[of=Aplus and Aminus];
                    \addplot[red!80, fill opacity=0.1] fill between[of=Bplus and Bminus];
                    \addplot[green!80, fill opacity=0.1] fill between[of=Cplus and Cminus];
\end{axis}

\end{tikzpicture}}
  \caption{
    \label{fig:experiment_1b}
    Average optimal value for experiment (1B) when the initial mean is far from the global minimum with a wide covariance.
  }
\end{figure}

\begin{figure}[!h]
  \centering
  \resizebox{0.7\columnwidth}{!}{\input{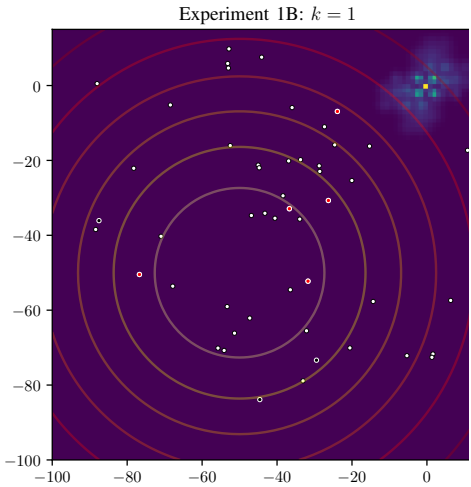}}
  \caption{
    \label{fig:example_1b}
    First iteration of the scenario in experiment (1B) where the initial distribution is far away form the global optimal. The red dots indicate the true-elites, the black dots with white outlines indicate the ``non-elites'' evaluated from the true objective function, and the white dots with black outlines indicate the samples evaluated using the surrogate model.
  }
\end{figure}

When the initial mean of the input distribution is placed far away from the global optimal, the CE-method tends to converge prematurely as shown in \cref{fig:experiment_1b}.
This scenario is illustrated in \cref{fig:example_1b}.
We can see that both CE-surrogate and CE-mixture perform well in this case.

\begin{figure}[!ht]
  % \centering
  \resizebox{0.9\columnwidth}{!}{\begin{tikzpicture}[]
\begin{axis}[height = {6cm}, legend style = {{at={(0.01,0.01)},anchor=south west}}, ylabel = {$\bar{b}_v$}, title = {Experiment 1C}, xmin = {1}, xmax = {10}, xlabel = {Iteration}, width = {10cm}]\addplot+ [mark = {none}, blue]coordinates {
(1.0, -0.002317341737835217)
(2.0, -0.004409663148327077)
(3.0, -0.005597712634610402)
(4.0, -0.005924407326871117)
(5.0, -0.006240681168005196)
(6.0, -0.006336869777484001)
(7.0, -0.00645335964644521)
(8.0, -0.006498832536568549)
(9.0, -0.006512999909560328)
(10.0, -0.006515684357317801)
};
\addlegendentry{CE-method}
\addplot+ [mark = {none}, red]coordinates {
(1.0, -0.0032523297583967304)
(2.0, -0.005055935736024911)
(3.0, -0.007254343579734129)
(4.0, -0.009425793030696876)
(5.0, -0.010381223749346389)
(6.0, -0.011697303259215135)
(7.0, -0.012359244989274664)
(8.0, -0.013041466197121826)
(9.0, -0.015631139919847138)
};
\addlegendentry{CE-surrogate}
\addplot+ [mark = {none}, green!50!black]coordinates {
(1.0, -0.003718899914679905)
(2.0, -0.005975318324663225)
(3.0, -0.008091989292448459)
(4.0, -0.009279493682692748)
(5.0, -0.010588537390066141)
(6.0, -0.011487546373826613)
(7.0, -0.012119978015104808)
(8.0, -0.012484799647570616)
(9.0, -0.013261146870849947)
(10.0, -0.014645709933698618)
};
\addlegendentry{CE-mixture}
\addplot+ [mark = {none}, dashed, blue, name path=Aplus, opacity=0.2]coordinates {
(1.0, -2.505055686322869e-5)
(2.0, -0.0006614193413984025)
(3.0, -0.0011266247205687892)
(4.0, -0.0013803909598864546)
(5.0, -0.001658432456737093)
(6.0, -0.0017166511411978603)
(7.0, -0.0017967264523970272)
(8.0, -0.001838149454357168)
(9.0, -0.0018469228764020481)
(10.0, -0.0018487907766789472)
};
\addplot+ [mark = {none}, dashed, blue, name path=Aminus, opacity=0.2]coordinates {
(1.0, -0.0046096329188072055)
(2.0, -0.00815790695525575)
(3.0, -0.010068800548652015)
(4.0, -0.01046842369385578)
(5.0, -0.010822929879273298)
(6.0, -0.010957088413770142)
(7.0, -0.011109992840493393)
(8.0, -0.011159515618779928)
(9.0, -0.011179076942718608)
(10.0, -0.011182577937956656)
};
\addplot+ [mark = {none}, dashed, red, name path=Bplus, opacity=0.2]coordinates {
(1.0, -0.0005911030782760943)
(2.0, -0.001976609661316075)
(3.0, -0.0031050508422256345)
(4.0, -0.0051624457056261884)
(5.0, -0.006323763373178248)
(6.0, -0.007750121884409469)
(7.0, -0.008445510156223423)
(8.0, -0.00921161194337762)
(9.0, -0.012753813024120047)
};
\addplot+ [mark = {none}, dashed, red, name path=Bminus, opacity=0.2]coordinates {
(1.0, -0.005913556438517367)
(2.0, -0.008135261810733747)
(3.0, -0.011403636317242623)
(4.0, -0.013689140355767564)
(5.0, -0.01443868412551453)
(6.0, -0.0156444846340208)
(7.0, -0.016272979822325905)
(8.0, -0.01687132045086603)
(9.0, -0.01850846681557423)
};
\addplot+ [mark = {none}, dashed, green!50!black, name path=Cplus, opacity=0.2]coordinates {
(1.0, -0.0006097905928366437)
(2.0, -0.0023709995253624262)
(3.0, -0.004127557677415865)
(4.0, -0.0053789760978201805)
(5.0, -0.006856131057669737)
(6.0, -0.007556847315585175)
(7.0, -0.008035559101084024)
(8.0, -0.008303268643850135)
(9.0, -0.00918373936480171)
(10.0, -0.009792935462045668)
};
\addplot+ [mark = {none}, dashed, green!50!black, name path=Cminus, opacity=0.2]coordinates {
(1.0, -0.006828009236523167)
(2.0, -0.009579637123964025)
(3.0, -0.012056420907481052)
(4.0, -0.013180011267565316)
(5.0, -0.014320943722462546)
(6.0, -0.015418245432068052)
(7.0, -0.016204396929125592)
(8.0, -0.016666330651291097)
(9.0, -0.017338554376898185)
(10.0, -0.019498484405351568)
};
\addplot[blue!80, fill opacity=0.1] fill between[of=Aplus and Aminus];
                    \addplot[red!80, fill opacity=0.1] fill between[of=Bplus and Bminus];
                    \addplot[green!80, fill opacity=0.1] fill between[of=Cplus and Cminus];
\end{axis}

\end{tikzpicture}}
  \caption{
    \label{fig:experiment_1c}
    Average optimal value for experiment (1C) when we restrict the number of objective function calls.
  }
\end{figure}
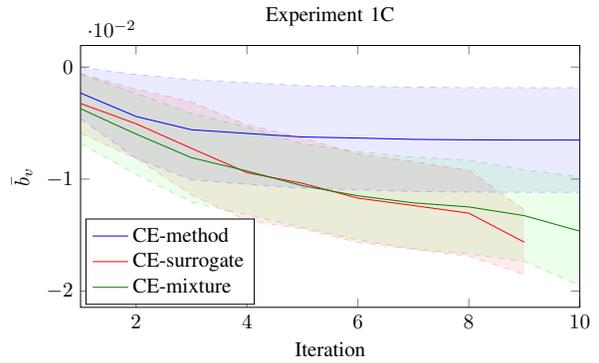

Given the same centered mean as before, when we restrict the number of objective function calls even further to just 50 we see interesting behavior.
Notice that the results of experiment (1C) shown in \cref{fig:experiment_1c} follow a curve closer to the far away mean from experiment (1B) than from the same setup as experiment (1A). Also notice that the CE-surrogate results cap out at iteration 9 due to the evaluation schedule front-loading the objective function calls, thus leaving none for the final iteration (while still maintaining the same total number of evaluations of 50).

\section{Conclusion} \label{sec:conclusion}
We presented variants of the popular cross-entropy method for optimization of objective functions with multiple local minima.
Using a Gaussian processes-based surrogate model, we can use the same number of true objective function evaluations and achieve better performance than the standard CE-method on average.
We also explored the use of a Gaussian mixture model to help find global minimum in multimodal objective functions.
We introduce a parameterized test objective function with a controllable global minimum and spread of local minima.
Using this test function, we showed that the CE-surrogate algorithm achieves the best performance relative to the standard CE-method, each using the same number of true objective function evaluations.

\printbibliography

\end{document}